\documentclass{article} 
\usepackage{iclr2016_conference,times}
\usepackage{amsfonts, amssymb, amsmath}
\usepackage{algorithm, algorithmic, amsthm}
\usepackage{graphicx}
\usepackage{color}
\usepackage{hyperref}
\usepackage[top=2.5cm, bottom=3cm, left=3cm, right=3cm]{geometry}

\usepackage{subfigure}

\usepackage{hyperref}
\usepackage{url}
\usepackage{graphicx, wrapfig}
\usepackage{amsfonts, amssymb, amsmath}
\usepackage{algorithm, algorithmic, amsthm}
\usepackage{multirow}
\usepackage{sidecap}

\newcommand{\rr}{\ensuremath{\mathbf{r}}}
\newcommand{\sss}{\ensuremath{\mathbf{s}}}  

\newcommand{\vv}{\ensuremath{\mathbf{v}}}

\newcommand{\x}{\ensuremath{\mathbf{x}}}

\newcommand{\calM}{\ensuremath{\mathcal{M}}}

{%
\begin{list}{#1}{
\vspace{-\topsep}
\vspace{-\partopsep}
\setlength{\itemindent}{0cm}
\setlength{\rightmargin}{0cm}
\setlength{\listparindent}{0cm}
\settowidth{\labelwidth}{#1}
\setlength{\leftmargin}{\labelwidth}
\addtolength{\leftmargin}{\labelsep}
\setlength{\itemsep}{0cm}
}%
}%
{%
\end{list}
\vspace{-\topsep}
\vspace{-\partopsep}
}

{\begin{enumerate}%
}%
{\end{enumerate}}

\theoremstyle{plain}

\newtheorem*{lemma*}{Lemma}

\newtheorem*{prop*}{Proposition}

\theoremstyle{definition}

\newtheorem*{defn*}{Definition}

\newtheorem*{exmp*}{Example}

\newtheorem*{conj*}{Conjecture}

\theoremstyle{remark}

\newtheorem*{rmk*}{Remark}

\title{A Deep Memory-based  Architecture for Sequence-to-Sequence Learning}

\author{Fandong Meng$^1$\thanks{The work was done when the first author worked as intern at Noah's Ark Lab, Huawei Technologies.} \ \ Zhengdong Lu$^2$ \ Zhaopeng Tu$^2$ Hang Li$^2$ \ and Qun Liu$^{1}$ \\
        $^1$Institute of Computing Technology, Chinese Academy of Sciences\\
        {\tt \{mengfandong, liuqun\}@ict.ac.cn}\\
        $^2$Noah's Ark Lab, Huawei Technologies\\
        {\tt \{Lu.Zhengdong, Tu.Zhaopeng, HangLi.HL\}@huawei.com}
}

\begin{document}

\maketitle

\begin{abstract}
\noindent We propose \textsc{DeepMemory}, a novel deep architecture for sequence-to-sequence learning, which performs the task through a series of nonlinear transformations from the representation of the input sequence (e.g., a Chinese sentence) to the final output sequence (e.g., translation to English). Inspired by the recently proposed Neural Turing Machine~\citep{ntmgraves2014neural}, we store the intermediate representations in stacked layers of memories, and use read-write operations on the memories to realize the nonlinear transformations between the representations. The types of transformations are designed in advance but the parameters are learned from data. Through layer-by-layer transformations, \textsc{DeepMemory} can model complicated relations between sequences necessary for applications such as machine translation between distant languages. The architecture can be trained with normal back-propagation on sequence-to-sequence data, and the learning can  be easily scaled up to a large corpus. \textsc{DeepMemory} is broad enough to subsume the state-of-the-art neural translation model in~\citep{cho} as its special case, while significantly improving upon the model with its deeper architecture.  Remarkably, \textsc{DeepMemory}, being purely neural network-based, can achieve performance comparable to the traditional phrase-based machine translation system Moses with a small vocabulary and a modest parameter size.
\end{abstract}

\section{Introduction} 
\begin{figure}[h!]
  \begin{center}
  \hspace{-25pt}\textsf{\small Encoder-Decoder} \hspace{135pt} \textsf{\small Attention model} \\
  \includegraphics[width=0.4\textwidth]{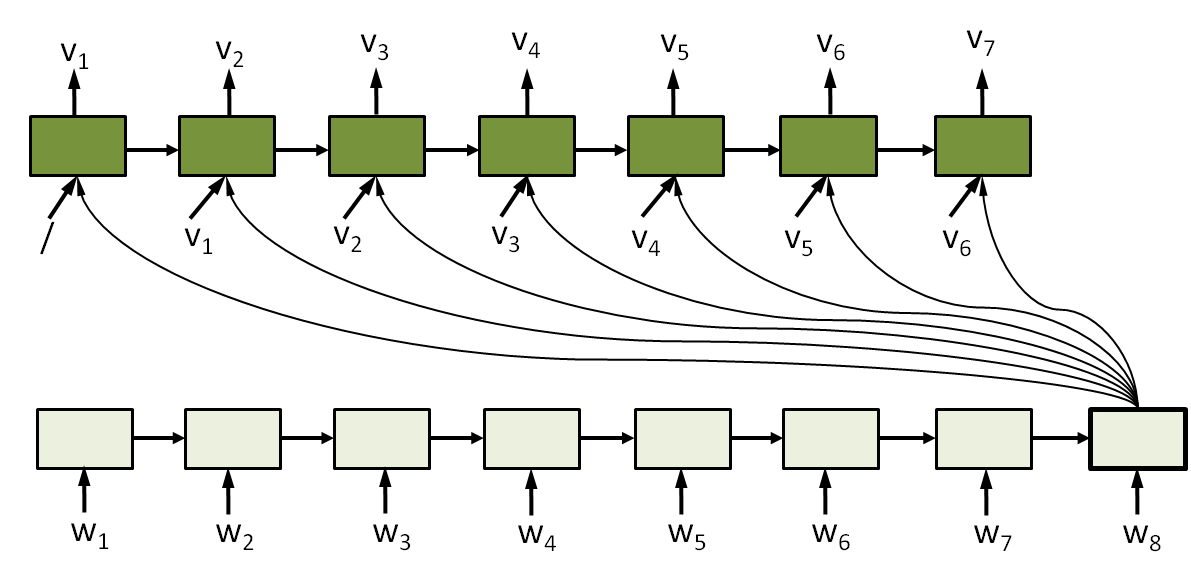} \hspace{30pt}
        \includegraphics[width=0.4\textwidth]{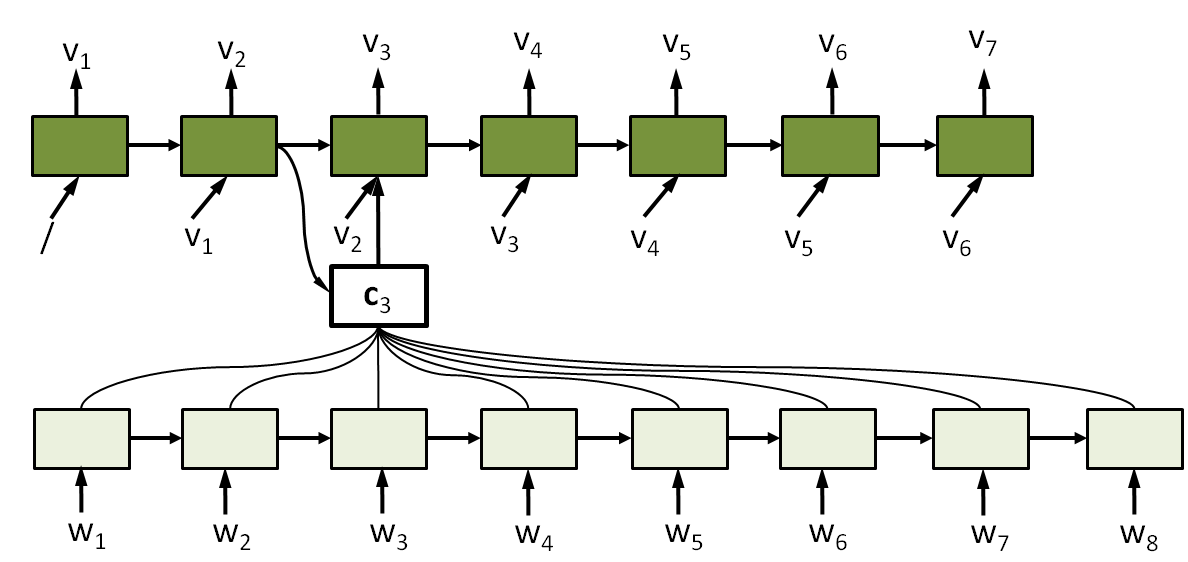}
          \vspace{-5pt}
   \caption{Two types of neural machine translators. Note the pictorial illustrations may deviate from individual models e.g., \cite{googleS2S}, on modeling details.  } 

    \label{f:2types}
  \end{center}\vspace{-8pt}
\end{figure}

\vspace{-5pt}
Sequence-to-sequence learning is a fundamental problem in natural language processing, with many important applications such as machine translation~\citep{cho,googleS2S}, part-of-speech tagging \citep{collobert2011natural,vinyals2014grammar} and dependency parsing \citep{chen2014fast}. Recently, there has been significant progress in development of technologies for the task using purely neural network-based models. Without loss of generality, we consider machine translation in this paper.
Previous efforts on neural machine translation generally fall into two categories: \vspace{-5pt}
\begin{itemize}
  \item {\bf Encoder-Decoder:} As illustrated in left panel of Figure~\ref{f:2types},  models of this type first summarize the source sentence into a fixed-length vector by the encoder, typically implemented with a recurrent neural network (RNN) or a convolutional neural network (CNN), and then unfold the vector into the target sentence by the decoder, typically implemented with a RNN~\citep{auli2013,kalchbrenner2013,ChoEMNLP,googleS2S};

  \item {\bf Attention-Model:} with RNNsearch~\citep{cho,luongEMNLP2015} as representative, it represents the source sentence as a sequence of vectors after a RNN (e.g., a bi-directional RNN~\citep{schuster1997bidirectional}), and then simultaneously conducts dynamic alignment with a gating neural network and generation of the target sentence with another RNN, as illustrated in right panel of Figure~\ref{f:2types}.   \vspace{-4pt}
      \end{itemize}

Empirical comparison between the approaches indicates that the attention-model is more efficient than the encoder-decoder approach: it can achieve comparable results with far less parameters and training instances~\citep{jean-EtAl:2015:ACL-IJCNLP}. This superiority in efficiency comes mainly from the mechanism of dynamic alignment, which avoids the need to represent the entire source sentence with a fixed-length vector~\citep{googleS2S}.

\subsection{Deep Memory-based Architecture} \vspace{-6pt}

\begin{figure}[h!]
\begin{center}
        \includegraphics[width=0.32\textwidth]{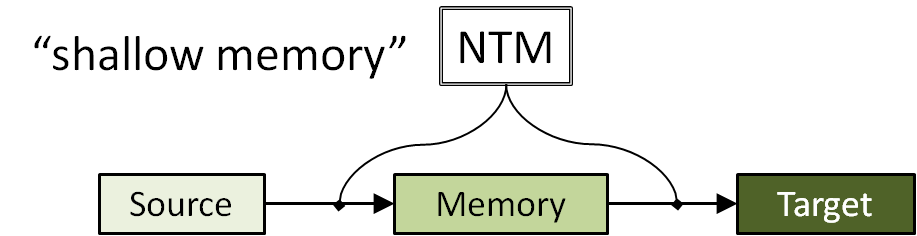}  \hspace{20pt}
        \includegraphics[width=0.62\textwidth]{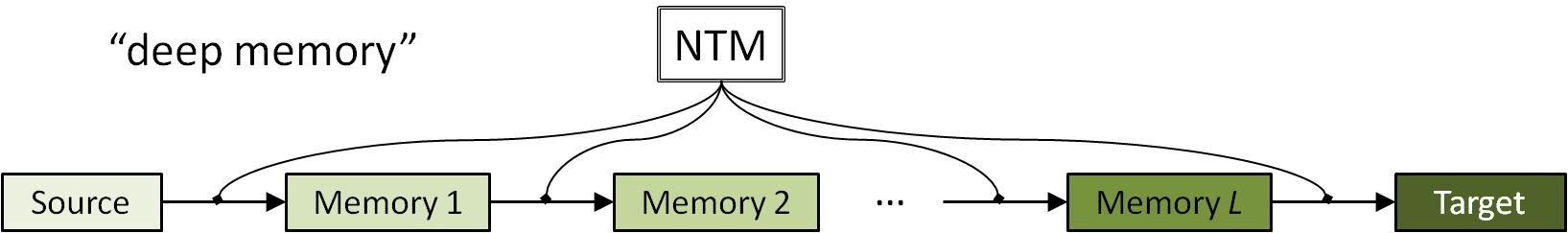}
   \caption{Space holder for NTM view of things} 
    \vspace{-12pt}
    \label{f:ntm_view}
  \end{center}
\end{figure}

Both encoder-decoders and attention models can be reformalized in the language of Neural Turing Machines (NTM)~\citep{ntmgraves2014neural}, by replacing different forms of representations as content in memories and the operations on them as basic neural net-controlled read-write actions, as illustrated in the left panel of Figure \ref{f:ntm_view}. This is clear after realizing that the attention mechanism~\citep{cho} is essentially a special case of reading (in particular with content-based addressing ) in NTM on the memory that contains the representation of source sentence.
More importantly, under this new view, the whole process becomes transforming the source sentence and putting it into memory (vector or array of vectors), and reading from this memory to further transform it into the target sentence. This architecture is intrinsically shallow in terms of the transformations on the sequence as an object, with essentially one hidden ``layer", as illustrated in the left panel of Figure \ref{f:ntm_view}. Note that although RNN (as encoding/decoding or equivalently as controller in NTM) can be infinitely deep, this depth is merely for dealing with the temporal structure \emph{within} the sequence. On the other hand, many sequence-to-sequence tasks, e.g, translation,  are intrinsically complex and calls for more complex and powerful transformation mechanism than that in encoder-decoder and attention models. \vspace{-1pt}

 For this reason, we propose a novel deep memory-based architecture, named \textsc{DeepMemory}, for sequence-to-sequence learning.  As shown in the right panel of Figure~\ref{f:ntm_view}, \textsc{DeepMemory} carries out the task through a series of non-linear transformations from the input sequence, to different levels of intermediate memory-based representations, and eventually to the final output sequence. \textsc{DeepMemory} is essentially a customized and deep version of NTM with multiple stages of operations controlled by a program, where the choices of ``layers" and types of read/write operations between layers are tailored for a particular task. 

%

Through layer-by-layer stacking of transformations on memory-based representations, \textsc{DeepMemory} generalizes the notion of inter-layer nonlinear mapping in neural networks, and therefore introduces a powerful new deep architecture for sequence-to-sequence learning.
The aim of \textsc{DeepMemory} is to learn the representation of sequence better suited to the task (e.g., machine translation) through layer-by-layer transformations. Just as in deep neural network (DNN), we expect that stacking relatively simple transformations can greatly enhance the expressing power and the efficiency of \textsc{DeepMemory}, especially in handling translation between languages with vastly different nature (e.g., Chinese and English) and sentences with complicated structures. \textsc{DeepMemory} naturally subsumes current neural machine translation models~\citep{cho,googleS2S} as special cases, but more importantly it accommodates many deeper alternatives with more modeling power, which are empirically superior to the current shallow architectures on machine translation tasks. 


Although \textsc{DeepMemory} is initially proposed for machine translation, it can be  adapted for other tasks that require substantial transformations of sequences, including paraphrasing, reasoning~\citep{Reasoning2015}, and semantic parsing~\citep{Enquirer2015}.  Also, in defining the layer-by-layer transformations, we can go beyond the read-write operations proposed in~\citep{ntmgraves2014neural} and design differentiable operations for the specific structures of the task (e.g., in ~\citep{Enquirer2015}).

\paragraph{RoadMap}
We will first discuss in Section \ref{s:rw} the read-write operations as a new form of non-linear transformation, as the building block of \textsc{DeepMemory}. Then in Section \ref{s:dm}, we stack the transformations together to get the full \textsc{DeepMemory} architecture, and discuss several architectural variations of it. In Section \ref{s:expts} we report our empirical study of \textsc{DeepMemory} on a Chinese-English translation task.

\section{Read-Write as a Nonlinear Transformation} \label{s:rw} \vspace{-5pt}

\begin{figure}[h!]
\begin{center}
              \includegraphics[width=0.44\textwidth]{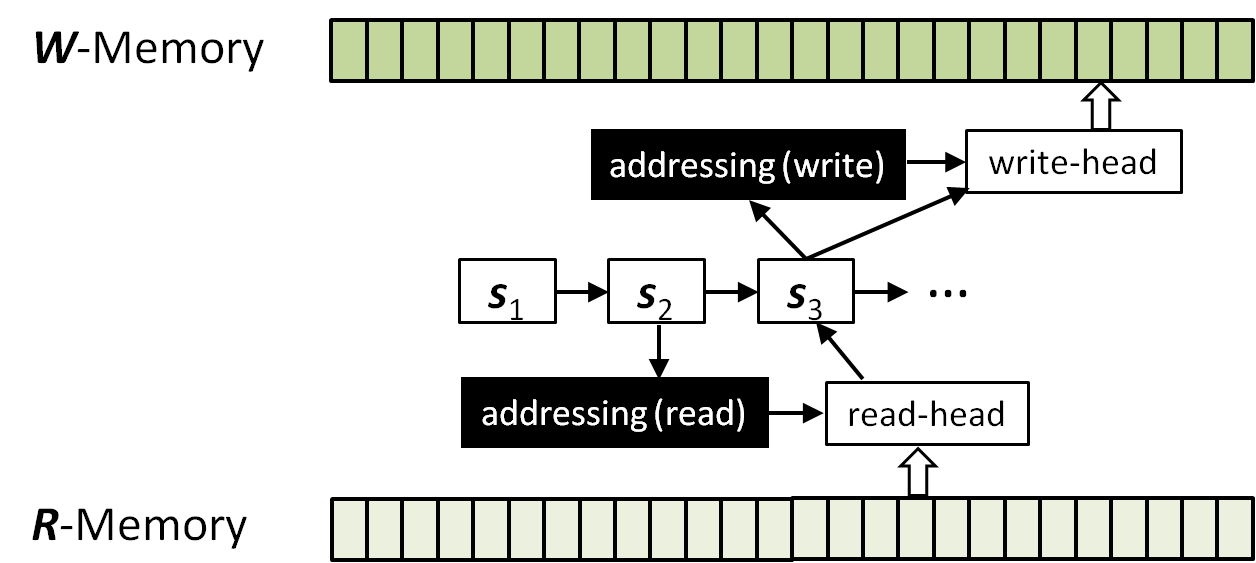} \hspace{70pt}
     \includegraphics[width=0.2\textwidth]{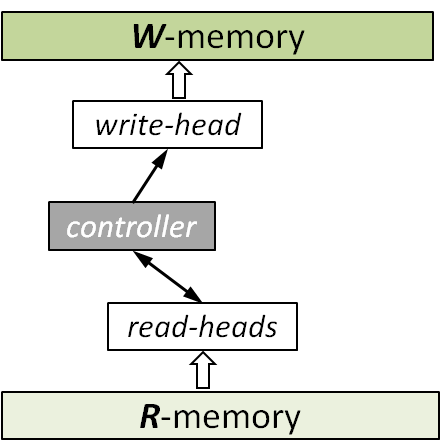}
   \caption{Read-write as an nonlinear transformation.} 
    \vspace{-12pt}
    \label{f:2layers}
  \end{center}
\end{figure}

We start with discussing read-write operations between two pieces of memory as a generalized form of nonlinear transformation. As illustrated in Figure~\ref{f:2layers} (left panel), this transformation is between two \emph{non-overlapping} memories, namely $R$-memory and $W$-memory, with $W$-memory being initially blank. A controller operates the read-heads to get the values from $R$-memory (``reading"), which are then sent to the write-head for modifying the values at specific locations in $W$-memory (``writing"). After those operations are completed, the content in $R$-memory is considered transformed and written to $W$-memory. These operations therefore define a transformation from one representation (in $R$-memory) to another (in $W$-memory), which is pictorially noted in the right panel of Figure \ref{f:2layers}.

These basic components are more formally defined below, following those in a generic NTM~\citep{ntmgraves2014neural}, with however important modifications for the nesting architecture, implementation efficiency and description simplicity. \vspace{-4pt}

\paragraph{Memory:} a memory is generally defined as a matrix with potentially infinite size, while here we limit ourselves to pre-determined (pre-claimed) $N\times d$ matrix, with $N$ memory locations and $d$ values in each location. In our implementation of \textsc{DeepMemory}, $N$ is always instance-dependent and is pre-determined by the algorithm\footnote{It is possible to let the controller learn to determine the length of the memory, but that does not yield better performance on our tasks and is therefore omitted here.}. Memories of different layers generally have different $d$. Now suppose for one particular instance (index omitted for notational simplicity), the system reads from the $R$-memory ($\calM^\textsc{r}$, with $N_\textsc{r}$ units) and writes to $W$-memory (denoted $\calM^\textsc{w}$, with $N_\textsc{w}$ units) \vspace{+2pt}
\[
R\text{-memory:}\;\;\calM^\textsc{r} = \{\x_{1}^\textsc{r},\;\x_{2}^\textsc{r},\;\cdots,\; \x_{N_\textsc{r}}^\textsc{r}\},\hspace{10pt}W\text{-memory:}\;\;\calM^\textsc{w} = \{\x_{1}^\textsc{w},\;\x_{2}^\textsc{w},\;\cdots,\; \x_{N_\textsc{w}}^\textsc{w}\},
\]
with $\x_{n}^\textsc{r} \in \mathbb{R}^{d_\textsc{r}}$ and $\x_{n}^\textsc{w} \in \mathbb{R}^{d_\textsc{w}}$.

\paragraph{Read/write heads:} a read-head gets the values from the corresponding memory, following the instructions of the controller, which also influences the {controller} in feeding the state-machine. \textsc{DeepMemory} allows multiple read-heads for one controller, with potentially different addressing strategies (see Section~\ref{s:Address} for more details). A write-head simply takes the instruction from {controller} and modifies the values  at specific locations.

\paragraph{Controller:} The core to the controller is a state machine, implemented as a RNN  Long Short-Term Memory (LSTM)~\citep{lstm}, with state at time $t$ denoted as $\sss_t$ (as illustrated in Figure~\ref{f:2layers}). With $\sss_t$, the controller determines the reading and writing at time $t$, while the return of reading in turn takes part in updating the state. For simplicity, only one reading and writing is allowed at one time step, but more than one read-heads are allowed. The main equations for controllers are then \vspace{+2pt}
\begin{eqnarray*}
&&\textsc{Read vector:}\hspace{16pt}\rr_{t} =  F_\textsc{r}(\calM^\textsc{r}, \sss_t; \;\Theta_{\textsc{r}})\\
&&\textsc{Write vector:}\hspace{10pt}\vv_t = F_\textsc{w} (\sss_{t}; \; \Theta_\textsc{w})\\
&&\textsc{State update:}\hspace{15pt} \sss_{t+1} = F_\textsc{d}(\sss_t, \rr_t;\;\Theta_\textsc{d})
\end{eqnarray*}
where $F_\textsc{d}(\cdot)$, $F_\textsc{r}(\cdot)$ and $F_\textsc{w}(\cdot)$  are respectively the operators for dynamics, reading and writing\footnote{Note that our definition of writing is slightly different from that in~\citep{ntmgraves2014neural}.},  parameterized by $\Theta_\textsc{d}$, $\Theta_\textsc{r}$, and $\Theta_\textsc{w}$. In \textsc{DeepMemory}, 1) it is only allowed to read from memory of lower layer and write to memory of higher layer, and 2) reading memory can only be performed after finishing the writing to it. 

The above read-write operations transform the representation in $R$-memory to the new representation in $W$-memory, while the design choice of read-write specifies the inner structure of the memories in  $W$-memory.  The transformation is therefore jointly specified by the read-write strategies (e.g., the addressing described in Section \ref{s:Address}) and the parameters learned in a supervised fashion (described later in Section~\ref{s:opt}). Memory with the designed inner structure, in this particular case vector array with instance specific length, offers more flexibility than a fixed-length vector in representing sequences. This representational flexibility is particularly advantageous  when combined with proper reading-writing strategy in defining nonlinear transformations for sequences, which will serve as the building block of the deep architecture.

\subsection{Addressing}\label{s:Address}
\subsubsection{Addressing for Reading}  \vspace{-5pt}
\paragraph{Location-based Addressing}
With location-based addressing ($L$-addressing), the reading is simply
$
\rr_{t} =  \x_{t}^\textsc{r}.
$
Notice that with $L$-addressing, the state machine automatically runs on a clock determined by the spatial structure of $R$-memory. Following this clock, the write-head operates the same number of times. One important variant, as suggested in~\citep{cho,googleS2S}, is to go through $R$-memory backwards after the forward reading pass, where the controller RNN has the same structure but is parameterized differently.

\paragraph{Content-based Addressing}
With a content-based addressing ($C$-addressing), the return at $t$ is
\begin{eqnarray*}
\rr_{t} = F_\textsc{r}(\calM^\textsc{r}, \sss_t; \Theta_{\textsc{r}})= \sum_{n=1}^{N_\textsc{r}} \tilde{g}(\sss_t, \x_{n}^\textsc{r} ; \Theta_{\textsc{r}}) \x_{n}^\textsc{r}, \;\; \text{ and }\;\;\tilde{g}(\sss_t, \x_{n}^\textsc{r} ; \Theta_{\textsc{r}}) = \frac{g(\sss_t, \x_{n}^\textsc{r} ; \Theta_{\textsc{r}})}{\sum_{n'=1}^{N_\textsc{r}}g(\sss_t, \x_{n'}^\textsc{r} ; \Theta_{\textsc{r}})},
\end{eqnarray*}
where $g(\sss_t, \x_{n}^\textsc{r} ; \Theta_{\textsc{r}})$, implemented as a DNN, gives an un-normalized ``affliiation" score for unit $\x^\textsc{r}_n$ in $R$-memory.  Clearly it is related to the attention mechanism introduced in~\citep{cho} for machine translation and general attention models discussed in~\citep{deepmind} for computer vision. Content-based addressing offers the following two advantages in representation learning:\vspace{-7pt}
\begin{enumerate}
  \item it can focus on the right segment of the representation, as demonstrated by the automatic alignment observed in~\citep{cho}, therefore better preserving the information in lower layers; \vspace{-2pt}
  \item it provides a way to alter the spatial structure of the sequence representation on a large scale, for which the re-ordering in machine translation is an intuitive example.
\end{enumerate}

\paragraph{Hybrid Addressing:}
With hybrid addressing ($H$-addressing) for reading, we essentially use two read-heads (can be easily extended to more), one with $L$-addressing
and the other with $C$-addressing. At each time $t$, the controller simply concatenates the return of two individual read-heads as the final return: \vspace{-5pt}
\[
\rr_{t} =  [\x_{t}^\textsc{r},\;\;\,\sum_{n=1}^{N_\textsc{r}} g(\sss_t, \x_{n}^\textsc{r} ; \Theta_{\textsc{r}}) \x_{n}^\textsc{r}]. \vspace{-1pt}
\]
It is worth noting that with $H$-addressing, the tempo of the state machine will be determined by the $L$-addressing read-head, and therefore creates  $W$-memory of the same number of locations in writing. As shown later, $H$-addressing can be readily extended to allow read-heads to work on different memories.

\subsubsection{Addressing for Writing} \vspace{-2pt}
\paragraph{Location-based Addressing}
With $L$-addressing, the writing is simple. At any time $t$, only the $t^{th}$ location in $W$-memory is updated:
$
\x^\textsc{W}_{t} =  \vv_t \overset{\text{def}}{=} F_\textsc{w} (\sss_{t}; \; \Theta_\textsc{w}),
$
which will be kept unchanged afterwards. For both location-  and content-based addressing, $F_\textsc{w}(\sss_t; \Theta_\textsc{w})$ is implemented as a DNN with weights $\Theta_\textsc{w}$.
\paragraph{Content-based Addressing}
In a way similar to $C$-addressing for reading, the units to write is determined through a gating network $g(\sss_t, \x^\textsc{w}_{n,t}; \Theta_\textsc{w})$, where the values in $W$-memory at time $t$ is given by
\[
 \Delta_{n,t} = \tilde{g}(\sss_t, \x^\textsc{w}_{n,t}; \Theta_\textsc{w})  F_\textsc{w}(\sss_t; \Theta_\textsc{w}), \;\;\; \x^\textsc{w}_{n,t} = (1-\alpha_t) \x^\textsc{w}_{n,t-1}+ \alpha_t \Delta_{n,t}, \; n = 1,2,\cdots, N_\textsc{w},
\]
where $\x^\textsc{w}_{n,t}$ stands for the values of the $n^{th}$ location in $W$-memory at time $t$, $\alpha$ is the forgetting factor (similarly defined as in~\citep{ntmgraves2014neural}), $ \tilde{g}$ is the normalized weight (with unnormalized score implemented also with a DNN) given to the $n^{th}$ location at time $t$.

\subsection{Types of Nonlinear Transformations} \label{s:sub_rw} \vspace{-5pt}

\begin{wrapfigure}{r}{0.4\textwidth}
\begin{center}
\vspace{-17pt}
    \begin{tabular}[c]{cc}
     \includegraphics[width=0.37\textwidth]{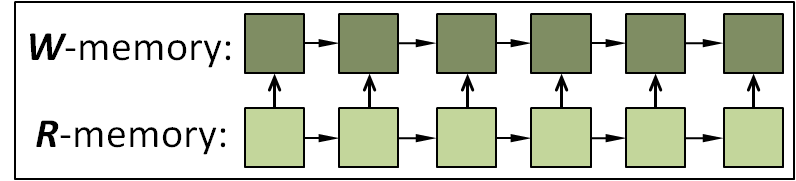}
     \end{tabular}
     \vspace{-20pt}
  \end{center}
\end{wrapfigure}
As the most ``conventional" special case, if we use $L$-addressing for both reading and writing, we  actually get the familiar structure in units found in RNN with stacked layers~\citep{pascanu2014construct}. Indeed, as illustrated in the figure right to the text, this read-write strategy will invoke a relatively local dependency based on the original spatial order in $R$-memory.  It is not hard to show that we can recover some deep RNN model in~\citep{pascanu2014construct} after stacking layers of read-write operations like this. This deep architecture actually partially accounts for the great performance of the Google neural machine translation model~\citep{googleS2S}.

\begin{figure}[h!]
\begin{center}
            \includegraphics[width=0.2\textwidth]{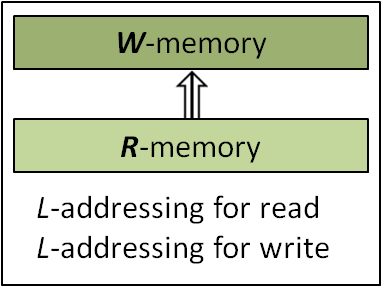}\hspace{5pt}
            \includegraphics[width=0.2\textwidth]{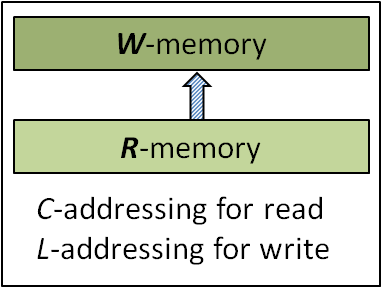}\hspace{5pt}
            \includegraphics[width=0.2\textwidth]{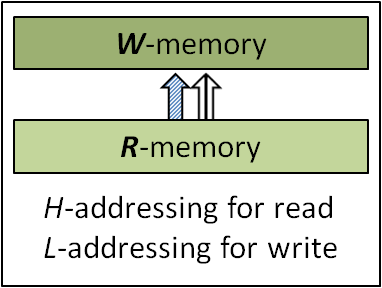}\hspace{5pt}
            \includegraphics[width=0.2\textwidth]{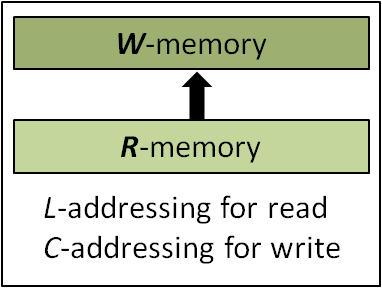}
      \caption{Examples of read-write strategies.}
    \label{f:AddressingTypes}
  \end{center}
\end{figure}

The $C$-addressing, however, be it for reading and writing, offers a means of major reordering on the units, while $H$-addressing can add into it the spatial structure of the lower layer memory. In this paper, we consider four types of transformations induced by combinations of the read and write addressing strategies, listed pictorially in Figure~\ref{f:AddressingTypes}. Notice that 1) we only include one combination with $C$-addressing for writing since it is computationally expensive to optimize when combined with a $C$-addressing reading (see Section~\ref{s:opt} for some analysis) , and 2) for one particular read-write strategy there are still a fair amount of implementation details to be specified, which are omitted due to the space limit. One can easily design different read/write strategies, for example a particular way of $H$-addressing for writing. \vspace{-5pt}

 \vspace{-10pt}

\section{\textsc{DeepMemory}: Stacking Them Together} \label{s:dm} \vspace{-5pt}

\begin{figure}[h!]
\begin{center}
      \hspace{20pt}\includegraphics[width=0.22\textwidth]{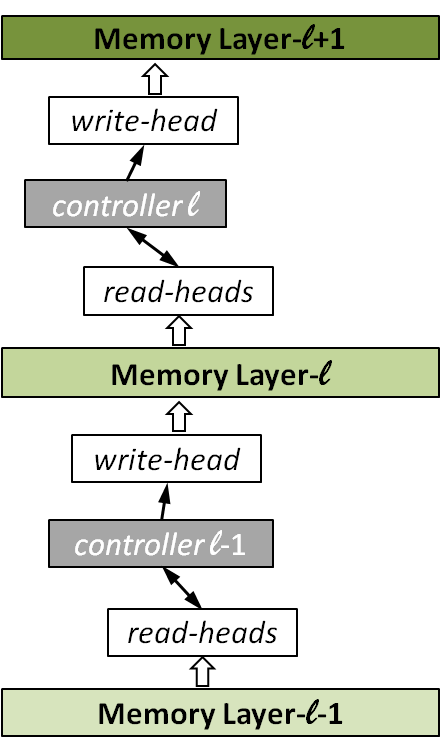} \hspace{0pt} \hspace{30pt}\includegraphics[width=0.24\textwidth]{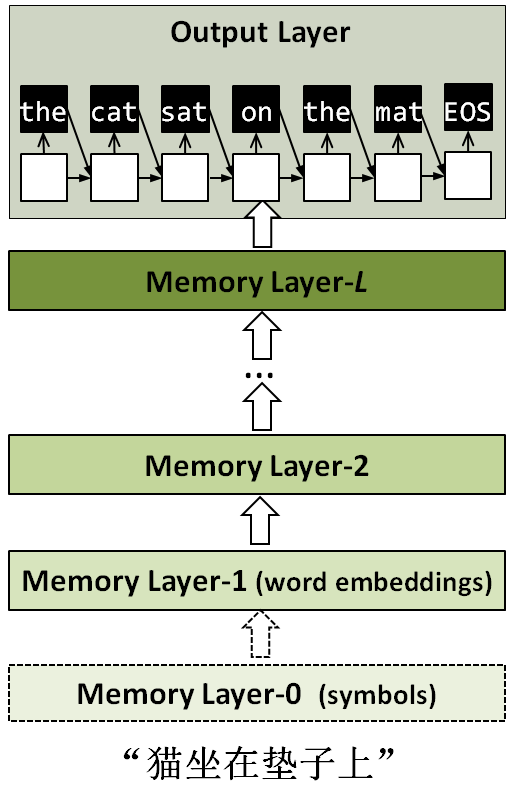}\vspace{-5pt}
      \caption{Illustration of stacked layers of memory (left) and the overall diagram of \textsc{DeepMemory} (right).}
    \label{f:DeepMemory}
  \end{center} \vspace{-10pt}
\end{figure}

As illustrated in Figure~\ref{f:DeepMemory} (left panel), the stacking is straightforward: we can just apply a transformation on top of another, with the $W$-memory in lower layer being the $R$-memory of upper layer. The entire deep architecture of \textsc{DeepMemory}, with diagram in Figure~\ref{f:DeepMemory} (right panel), can be therefore defined accordingly. Basically, it starts with a symbol sequence (Layer-0), then moves to the sequence of word embeddings (Layer-1), through layers of transformation to reach the final intermeidate layer (Layer-$L$), which will be read by the output layer.

The operations in output layer, relying on another LSTM to generate the target sequence, are similar to a memory read-write, with the following two differences: \vspace{-4pt}
\begin{itemize}
  \item it predicts the symbols for the target sequence, and takes the ``guess" as part of the input to update the state of the generating LSTM, while in a memory read-write, there is no information flow from higher layers to the controller; \vspace{-1pt}
  \item since the target sequence in general has different length as the top-layer memory, it takes only pure $C$-addressing reading and relies on the built-in mechanism of the generating LSTM to stop (i.e., after generating a \texttt{End-of-Sentence} token). \vspace{-2pt}
\end{itemize}

Memory of different layers could be equipped with different read-write strategies, and even for the same strategy, the configurations and learned parameters are in general different. This is in contrast to DNNs, for which the transformations of different layers are more homogeneous (mostly linear transforms with nonlinear activation function). A sensible architecture design in combining the nonlinear transformations can greatly affect the performance of the model, on which however little is known and future research is needed.

\subsection{Cross-Layer Reading}  \label{s:cross} \vspace{-5pt}
In addition to the generic read-write strategies in Section~\ref{s:sub_rw}, we also introduce the cross-layer reading into \textsc{DeepMemory} for more modeling flexibility. In other words, for writing in any Layer-$\ell$, \textsc{DeepMemory} allows reading from more than one layers lower than $\ell$, instead of just Layer-$\ell\hspace{-2pt}-\hspace{-2pt}1$. More specifically, we consider the following two cases. \vspace{-5pt}

\begin{wrapfigure}{r}{0.45\textwidth}
\begin{center}
\vspace{-10pt}
    \begin{tabular}[c]{cc}
     \includegraphics[width=0.45\textwidth]{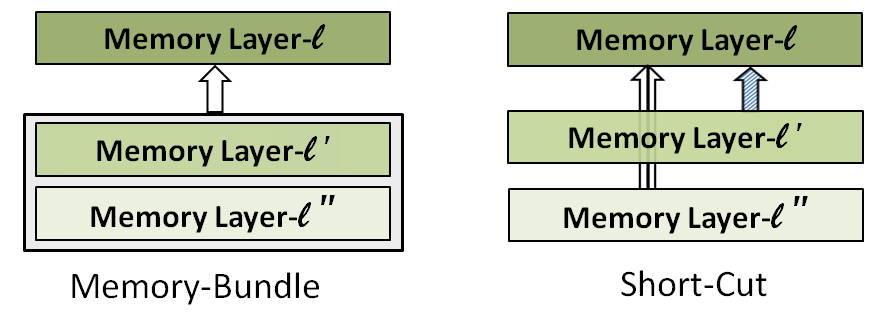}
     \end{tabular}
     \vspace{-15pt}
    \caption{Cross-layer reading.} \vspace{-5pt}
    \label{f:crosslayer}
  \end{center}  \vspace{-15pt}
\end{wrapfigure}


\paragraph{\textsc{Memory-Bundle:}} \label{s:crosslayer}
A \textsc{Memory-Bundle}, as shown in Figure~\ref{f:crosslayer} (left panel), concatenates the units of  two \emph{aligned} memories in reading, regardless of the addressing strategy.
Formally, the $n^{th}$ location in the bundle of memory Layer-$\ell'$ and Layer-$\ell''$ would be
$
\x_n^{(\ell' + \ell'')} = [(\x_n^{\ell'})^\top, (\x_n^{\ell''})^\top]^\top.
$
Since it requires strict alignment between the memories to put together, \textsc{Memory-Bundle} is usually on layers created with spatial structure of same origin (see Section \ref{s:special} for examples). \vspace{-8pt}

\paragraph{\textsc{Short-Cut}}
Unlike \textsc{Memory-Bundle}, \textsc{Short-Cut} allows reading from layers with potentially different inner structures by using multiple read-heads, as shown in Figure~\ref{f:crosslayer} (right panel). For example, one can use a $C$-addressing read-head on memory Layer-$\ell'$ and a $L$-addressing read-head on Layer-$\ell''$ for the writing to memory Layer-$\ell$ with $\ell', \ell''<\ell$.

\subsection{Optimization}  \label{s:opt} \vspace{-5pt}
For any designed architecture, the parameters to be optimized include $\{\Theta_\textsc{d}, \Theta_\textsc{r}, \Theta_\textsc{w}\}$ for each controller, the parameters for the LSTM in the output layer, and the word-embeddings. Since the reading from each memory can only be done after the writing on it completes, the ``feed-forward" process can be described in two scales: 1) the flow from memory of lower layer to memory of higher layer, and 2) the forming of a memory at each layer controlled by the corresponding state machine. Accordingly in optimization, the flow of ``correction signal" also propagates at two scales: \vspace{-5pt}
\begin{itemize}
  \item On the ``cross-layer" scale: the signal starts with the output layer and propagates from higher layers to lower layers, until Layer-1 for the tuning of word embedding;
  \item On the ``within-layer" scale: the signal back-propagates through time (BPTT) controlled by the corresponding state-machine (LSTM). In optimization, there is a correction for each reading or writing on each location in a memory, making the $C$-addressing more expensive than $L$-addressing for it in general involves all locations in the memory at each time $t$. \vspace{-5pt}

\end{itemize}
The optimization can be done via the standard back-propagation (BP) aiming to maximize the likelihood of the target sequence. In practice, we use the standard stochastic gradient descent (SGD) and mini-batch (size 80) with learning rate controlled by AdaDelta~\citep{adadelta}.

\subsection{Architectural Variations of DeepMemory} \label{s:special}
We discuss four representative special cases of \textsc{DeepMemory}: \textsc{Arc-I, II, III} and \textsc{IV}, as novel deep architectures for machine translation. We also show that current neural machine translation models like RNNsearch can be described in the framework of \textsc{DeepMemory} as a relatively shallow case. \vspace{-10pt}

\begin{wrapfigure}{r}{0.45\textwidth}
\begin{center}
\vspace{-14pt}
    \begin{tabular}[c]{cc}
     \includegraphics[width=0.22\textwidth]{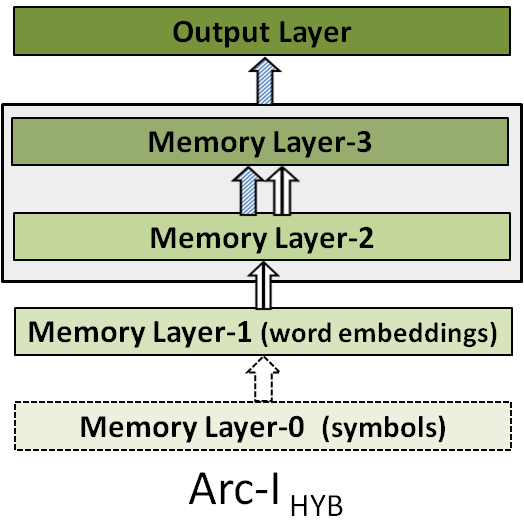}
     \includegraphics[width=0.22\textwidth]{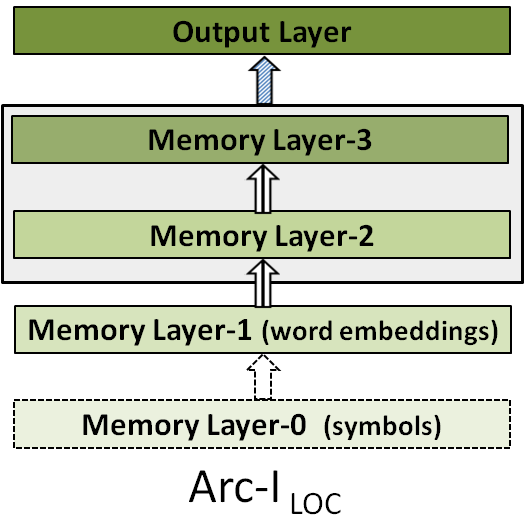}
     \end{tabular}
\vspace{-20pt}
  \end{center}
\end{wrapfigure}

\paragraph{\textsc{Arc-I}} The first proposal, including two variants (\textsc{Arc-I$_\textsc{hyb}$} and \textsc{Arc-I$_\textsc{loc}$}), is designed to demonstrate the effect of $C$-addressing reading between intermediate memory layers, with diagram shown in the figure right to the text. Both variants employ a $L$-addressing reading from memory Layer-1 (the embedding layer) and $L$-addressing writing to Layer-2. After that, \textsc{Arc-I$_\textsc{hyb}$} writes to Layer-3 ($L$-addressing) based on its $H$-addressing reading (two read-heads) on Layer-2, while \textsc{Arc-I$_\textsc{loc}$} uses $L$-addressing to read from Layer-2. Once Layer-3 is formed, it is then put together with Layer-2 for a \textsc{Memory-Bundle}, from which the output layer reads ($C$-addressing) for predicting the target sequence. \textsc{Memory-Bundle}, with its empirical advantage over single layers (see Section \ref{s:discuss}), is also used in other three architectures for generating the target sequence or forming intermediate layers.

\vspace{-10pt}


\begin{wrapfigure}{r}{0.27\textwidth}
\begin{center}
\vspace{-20pt}
    \begin{tabular}[c]{cc}
     \includegraphics[width=0.21\textwidth]{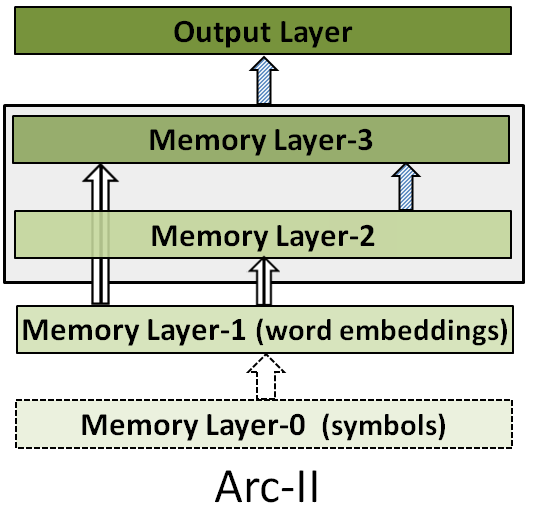}
     \end{tabular}
\vspace{-20pt}
  \end{center}
\end{wrapfigure}
\paragraph{\textsc{Arc-II}} As an architecture similar to \textsc{Arc-I}$_\textsc{hyb}$, \textsc{Arc-II} is designed to investigate the effect of $H$-addressing reading from different layers of memory (or \textsc{Short-Cut} in Section \ref{s:cross}). It uses the same strategy as \textsc{Arc-I}$_\textsc{hyb}$ in generating memory Layer-1 and 2, but differs in generating Layer-3, where \textsc{Arc-II} uses $C$-addressing reading on Layer-2 but $L$-addressing reading on Layer-1. Once Layer-3 is formed, it is then put together with Layer-2 as a \textsc{Memory-Bundle}, which is then read by the output layer for predicting the target sequence.

%

\begin{wrapfigure}{r}{0.27\textwidth}
\begin{center}
\vspace{-24pt}
    \begin{tabular}[c]{cc}
     \includegraphics[width=0.21\textwidth]{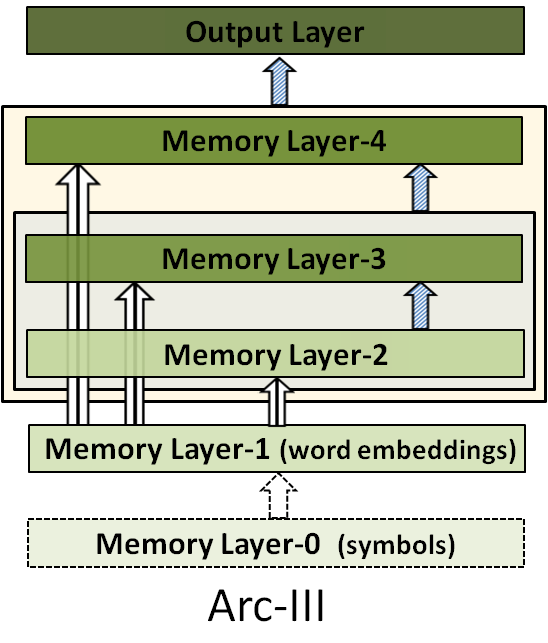}
     \end{tabular}
\vspace{-20pt}
  \end{center}
\end{wrapfigure}
\paragraph{\textsc{Arc-III}} We intend to use this design to study a deeper architecture and more complicated addressing strategy. \textsc{Arc-III} follows the same way  as \textsc{Arc-II} to generate Layer-1, Layer-2 and Layer-3. After that it uses two read-heads combined with a $L$-addressing write to generate Layer-4, where the two read-heads consist of a $L$-addressing read-head on Layer-1 and a $C$-addressing read-head on the memory bundle of Layer-2 and Layer-3. After the generation of Layer-4, it puts Layer-2, 3 and 4 together for a bigger \textsc{Memory-Bundle} to the output layer. \textsc{Arc-III}, with 4 intermediate layers, is the deepest among the four special cases.


\begin{wrapfigure}{r}{0.27\textwidth}
\begin{center}
\vspace{-20pt}
    \begin{tabular}[c]{cc}
     \includegraphics[width=0.21\textwidth]{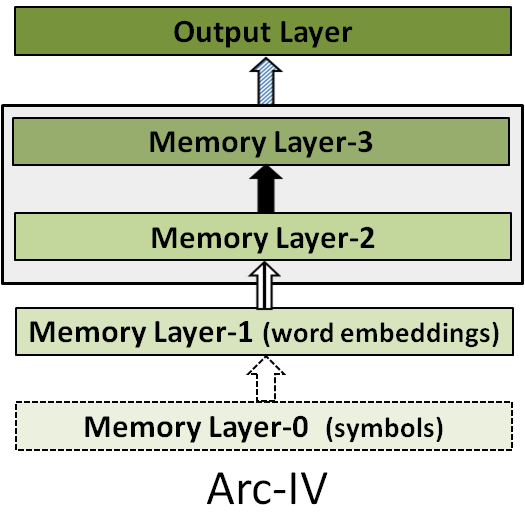}
     \end{tabular}
\vspace{-30pt}
  \end{center}
\end{wrapfigure}

\paragraph{Arc-IV} This proposal is designed to study the efficacy of $C$-addressing writing in forming intermediate representation. It employs a $L$-addressing reading from memory Layer-1 and $L$-addressing writing to Layer-2. After that, it uses a $L$-addressing reading on Layer-2 to write to Layer-3 with $C$-addressing. For  $C$-addressing writing to Layer-3,  all locations in Layer-3 are randomly initialized. Once Layer-3 is formed, it is then bundled with Layer-2 for the reading ($C$-addressing) of the output layer.



%

\subsubsection{Relation to other neural machine translators} 
\begin{figure}[h!]
\begin{center}
 \hspace{10pt}\textsf{\scriptsize Attention-model diagram} \hspace{225pt} \textsc{\scriptsize DeepMemory }\textsf{\scriptsize diagram} \\ \includegraphics[width=0.63\textwidth]{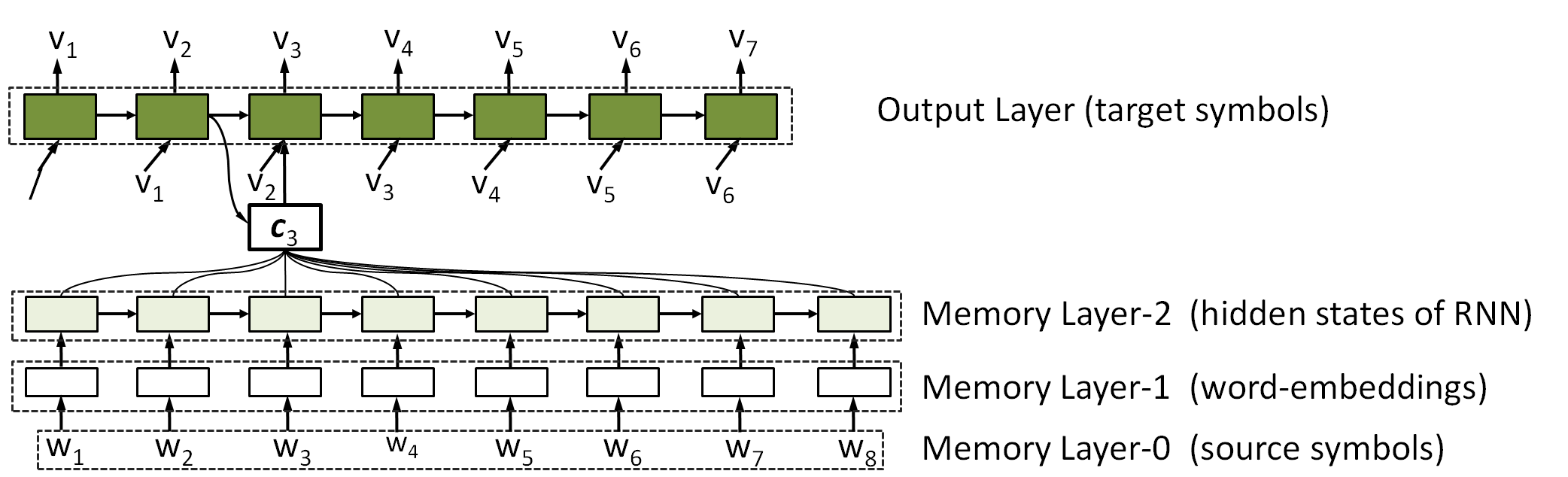} \hspace{38pt}
 \includegraphics[width=0.22\textwidth]{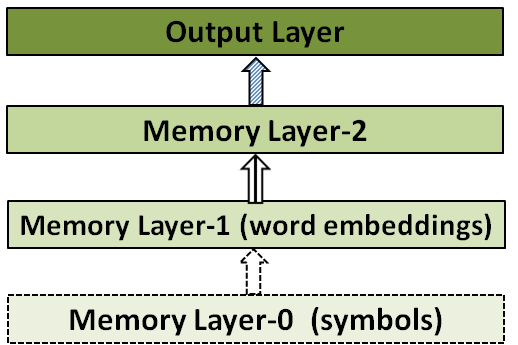}
     \caption{RNNsearch as a special case of \textsc{DeepMemory}.} \vspace{-15pt}
 \label{f:RNNsearch}
  \end{center}
\end{figure}

\vspace{-4pt}
As pointed out earlier, RNNsearch~\citep{cho} with its automatic alignment, is a special case of \textsc{DeepMemory} with shallow architecture. As pictorially illustrated in Figure~\ref{f:RNNsearch}, it employs $L$-addressing reading on memory Layer-1 (the embedding layer), and $L$-addressing writing to Layer-2, which then is read ($C$-addressing) by the output layer to generate the target sequence.  As shown in Figure~\ref{f:RNNsearch}, Layer-2 is the only intermediate layer created by nontrivial read-write operations. 

On the other hand, the connection between \textsc{DeepMemory} and encoder-decoder architectures is less obvious since they usually require the reading from only the last cell (i.e., for a fixed-length vector representation) between certain layers. More specifically, \cite{googleS2S} can be viewed as DeepMemory with stacking layers of $L$-addressing read-write (described in Section \ref{s:sub_rw}) for both the encoder and decoder part, while the two are actually connected through last hidden states of the LSTMs of the corresponding layers.

\section{Experiments} \label{s:expts} \vspace{-9pt}
We report our empirical study on applying \textsc{DeepMemory} to Chinese-to-English translation.
Our training data consist of 1.25M sentence pairs extracted from LDC corpora, with 27.9M Chinese words and 34.5M English words respectively.
We choose NIST 2002 (MT02) dataset as our development set, and the NIST 2003 (MT03), 2004 (MT04) and 2005 (MT05) datasets as our test sets.
We use the case-insensitive 4-gram NIST BLEU score as our evaluation metric, and \emph{sign-test}~\citep{collins2005clause} as statistical significance test.
In training of the neural networks, we limit the source and target vocabularies to the most frequent 16K words in Chinese and English, covering approximately 95.8\% and 98.3\% of the two corpora respectively.

We compare our method with two state-of-the-art SMT and NMT\footnote{There are recent progress on aggregating multiple models or enlarging the vocabulary(e.g., in ~\citep{jean-EtAl:2015:ACL-IJCNLP}), but here we focus on the generic models.} models:\vspace{-4pt}
\begin{itemize}
    \item {\bf Moses}~\citep{koehn2007}: an open source phrase-based translation system with default configuration and a 4-gram language model trained on the target portion of training data with~\citep{stolcke2002srilm};
  \item {\bf RNNsearch}~\citep{cho}:  an attention-based NMT model with default setting ({RNNsearch$_\textsc{default}$}), as well as an optimal re-scaling of the model (on sizes of both embedding and hidden layers, with about 50\% more parameters) ({RNNsearch$_\textsc{best}$}).
\end{itemize}  \vspace{-5pt}

For a fair comparison, 1) the output layer in each\textsc{DeepMemory} variant is implemented as Gated Recurrent Units (GRU) in~\citep{cho}, and 2) all the \textsc{DeepMemory} architectures are designed to have the same embedding size as in RNNsearch$_\textsc{default}$ with parameter size less or comparable to RNNsearch$_\textsc{best}$.

\subsection{Results} \label{s:result} \vspace{-6pt}
The main results of different models are given in Table~\ref{t:model-cpm}. RNNsearch (best) is about $1.7$ points behind Moses in BLEU
on average, which is consistent with the observations made by other authors on different machine translation tasks~\citep{cho,jean-EtAl:2015:ACL-IJCNLP}. Remarkably, some sensible designs of \textsc{DeepMemory} (e.g., \textsc{Arc-II}) can already achieve performance comparable to Moses, with only 42M parameters, while RNNsearch$_\textsc{best}$ has 46M parameters.   \vspace{-2pt}

Clearly all \textsc{DeepMemory} architectures yield performance significantly better than (\textsc{Arc-I}$_\textsc{hyb}$, \textsc{Arc-II} \& \textsc{Arc-III}) or comparable (\textsc{Arc-I}$_\textsc{loc}$ \& \textsc{Arc-IV}) to the NMT baselines. Among them, \textsc{Arc-II} outperforms the best setting of NMT baseline (RNNsearch$_\textsc{best}$), by about $1.5$ BLEU on average with less parameters. \vspace{-4pt}



\begin{table*}[h!]
\begin{center}
\scalebox{0.95}{
\begin{tabular}{l|llll|c}
\hline
\textsc{Systems} & \textsc{MT03} & \textsc{MT04} & \textsc{MT05} & \textsc{Average} & \textsc{Parameters} \# \\
\hline\hline
RNNsearch$_\textsc{default}$  				& 29.02    & 31.25    & 28.32    & 29.53  & 31M \\
RNNsearch$_\textsc{best}$  					& 30.28    & 31.72    & 28.52    & 30.17  & 46M \\

\hline\hline
\textsc{Arc-I}$_\textsc{loc}$  				& 28.98    & 32.02    & 29.53*    & 30.18  & 54M \\
\textsc{Arc-I}$_\textsc{hyb}$  				& 30.14    & 32.70*    & 29.40*    & 30.75  & 54M \\
\textsc{Arc-II}       						& \textbf{31.27}*    & 33.02*    & \textbf{30.63}*    & 31.64  & 42M \\
\textsc{Arc-III}        						& 30.15    & \textbf{33.46}*    & 29.49*    & 31.03  & 53M \\
\textsc{Arc-IV}         						& 29.88    & 32.00    & 28.76    & 30.21 & 48M \\
\hline\hline
Moses        							& 31.61    & 33.48    & 30.75    & 31.95  & -- \\
\hline
\end{tabular}
}
\end{center} \vspace{-4pt}
\caption{\label{t:model-cpm} BLEU-4 scores (\%) of NMT baselines: RNNsearch$_\textsc{default}$ and RNNsearch$_\textsc{best}$, \textsc{DeepMemory} architectures (\textsc{Arc-I, II,  III} and \textsc{IV}), and phrase-based SMT system (Moses).  The ``*" indicates that the results are significantly (p$<$0.05) better than those of the RNNsearch$_\textsc{best}$.
} \vspace{-10pt}
\end{table*}\vspace{-4pt}

\subsection{Discussion} \label{s:discuss} \vspace{-5pt}

\begin{wrapfigure}{r}{0.52\textwidth}
\begin{center}
\vspace{-10pt}
    \begin{tabular}[c]{cc}
   \includegraphics[width=0.5\textwidth]{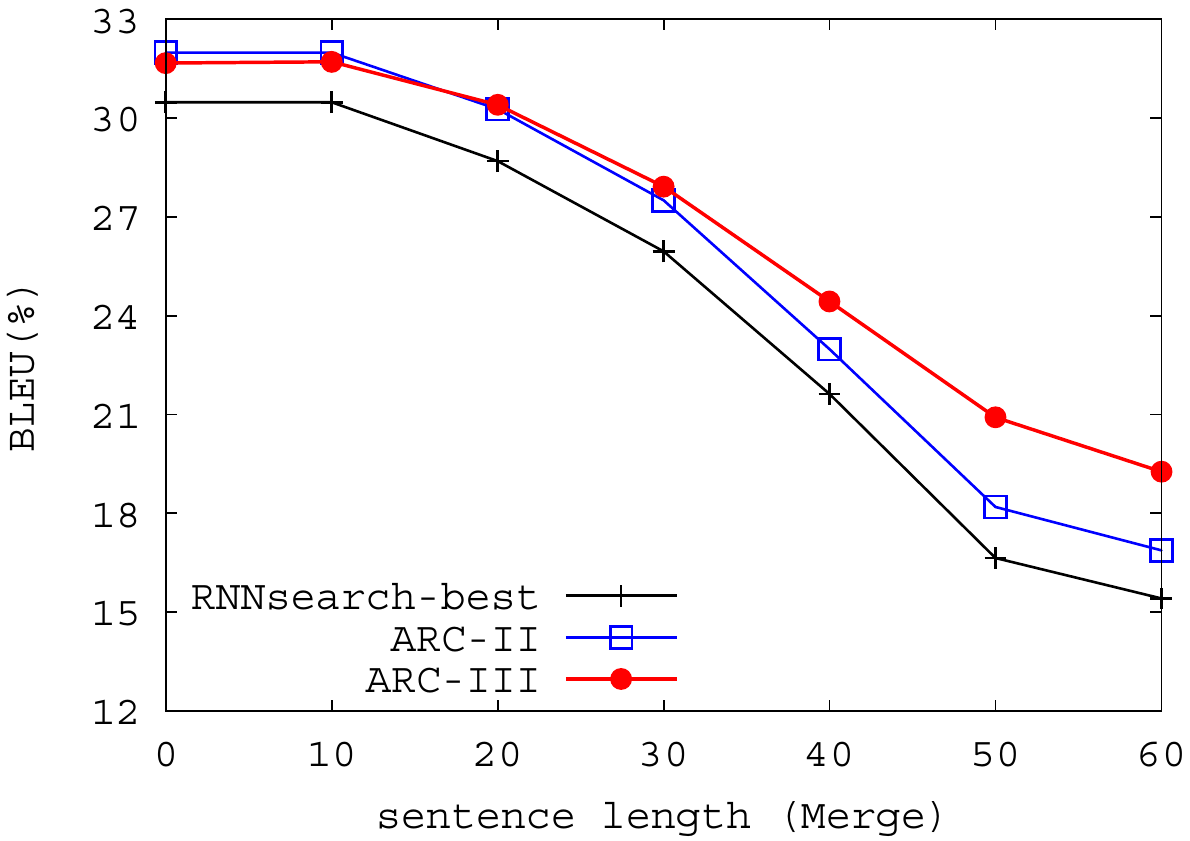}
     \end{tabular}
\vspace{-15pt}
  \end{center}
  \caption{The BLEU scores of generated translations on the merged three test sets with respect to the lengths of source sentences. The numbers on X-axis of the figure stand for sentences \emph{longer than} the corresponding length, e.g., $30$ for source sentences with $> 30$ words.}
\label{f:res_length}
\end{wrapfigure}

\paragraph{About Depth:} A more detailed comparison between RNNsearch (two layers), \textsc{Arc-II} (three layers)  and \textsc{Arc-III}) (four layers), both quantitatively and qualitatively, suggests that with deep architectures are essential to the superior performance of \textsc{DeepMemory}. Although the deepest architecture \textsc{Arc-III} is about 0.6 BLEU behind the \textsc{Arc-II}, its performance on long sentences is significant better.  Figure~\ref{f:res_length} shows the BLEU scores of generated translations on the test sets with respect to the length of the source sentences. In particular, we test the BLEU scores on sentences longer than \{0, 10, 20, 30, 40, 50, 60\} in the merged test set of MT03, MT04 and MT05. Clearly, on sentences with length $>$30, \textsc{Arc-III} yields consistently higher BLEU scores than \textsc{Arc-II}. This observation is further confirmed by our observations of translation quality (see Appendix) and is consistent with our intuition that \textsc{DeepMemory} with its multiple layers of transformation, is especially good at modeling the transformations of representations essential to machine translation on relatively complicated sentences. \vspace{-5pt}

\paragraph{About $C$-addressing Read:}Further comparison between \textsc{Arc-I}$_\textsc{hyb}$ and \textsc{Arc-I}$_\textsc{loc}$ (similar parameter sizes) suggests that  $C$-addressing reading plays an important role in learning a powerful transformation between intermediate representations, necessary for translation between language pairs with vastly different syntactical structures. This conjecture is further verified by the good performances of \textsc{Arc-II} and \textsc{Arc-III}, both of which have $C$-addressing read-heads in their intermediate memory layers. {\color{black} However, if memory Layer-$\ell\hspace{-3pt}+\hspace{-3pt}1$ is formed with only $C$-addressing read from memory Layer-$\ell$, and serves as the only going to later stages, the performance is usually less satisfying.} Comparison of this design with $H$-addressing (results omitted here) suggests that another read-head with $L$-addressing can prevent the transformation from going astray by adding the tempo from a memory with a clearer temporal structure. \vspace{-5pt}

\paragraph{About $C$-addressing Write:} The BLEU scores of \textsc{Arc-IV} are lower than that of \textsc{Arc-II} but comparable to that of RNNsearch$_\textsc{best}$, suggesting that writing with $C$-addressing alone yields reasonable representation. A closer look shows that although \textsc{Arc-IV} performs poorly on very long sentences (e.g., source sentences with over 60 words), it does fairly well on sentences with normal length. More specifically, on source sentences with length $\le$ 40, it outperforms RNNsearch$_\textsc{best}$  with 0.79 BLEU points. One possible explanation is that our particular implementation of $C$-addressing for writing in \textsc{Arc-IV} (Section \ref{s:special}) relies heavily on the randomly initialized content and is hard to optimize, especially when the structure of the sentence is complex, which might need to be ``guided" by another write-head or some smart initialization. \vspace{-5pt}

\paragraph{About Cross-layer Read:}As another observation, cross-layer reading almost always helps. The performances of \textsc{Arc-I, II,  III} and \textsc{IV} unanimously drop after removing the \textsc{Memory-Bundle} and \textsc{Short-Cut} (results omitted here), even after the broadening of memory units to keep the parameter size unchanged. It might be due to the flexibility gained in mixing different addressing modes and representations of different stages. \vspace{-8pt}

\section{Conclusion} \vspace{-8pt}
We propose \textsc{DeepMemory}, a novel architecture for sequence-to-sequence learning, which is stimulated by the recent work of Neural Turing Machine~\citep{ntmgraves2014neural} and Neural Machine Translation~\citep{cho}. \textsc{DeepMemory} builds its deep architecture for processing sequence data on the basis of a series of transformations induced by the read-write operations on a stack of memories. This new architecture significantly improves the expressive power of models in sequence-to-sequence learning, which is verified by our empirical study on a benchmark machine translation task.


\newpage
\bibliography{nips2015}

\begin{thebibliography}{21}
\providecommand{\natexlab}[1]{#1}
\providecommand{\url}[1]{\texttt{#1}}
\expandafter\ifx\csname urlstyle\endcsname\relax
  \providecommand{\doi}[1]{doi: #1}\else
  \providecommand{\doi}{doi: \begingroup \urlstyle{rm}\Url}\fi

\bibitem[Auli et~al.(2013)Auli, Galley, Quirk, and Zweig]{auli2013}
Auli, Michael, Galley, Michel, Quirk, Chris, and Zweig, Geoffrey.
\newblock Joint language and translation modeling with recurrent neural
  networks.
\newblock In \emph{Proceedings of EMNLP}, pp.\  1044--1054, 2013.

\bibitem[Bahdanau et~al.(2015)Bahdanau, Cho, and Bengio]{cho}
Bahdanau, Dzmitry, Cho, Kyunghyun, and Bengio, Yoshua.
\newblock Neural machine translation by jointly learning to align and
  translate.
\newblock In \emph{Proceedings of ICLR}, 2015.

\bibitem[Chen \& Manning(2014)Chen and Manning]{chen2014fast}
Chen, Danqi and Manning, Christopher~D.
\newblock A fast and accurate dependency parser using neural networks.
\newblock In \emph{Proceedings of EMNLP}, pp.\  740--750, 2014.

\bibitem[Cho et~al.(2014)Cho, van Merrienboer, Gulcehre, Bougares, Schwenk, and
  Bengio]{ChoEMNLP}
Cho, Kyunghyun, van Merrienboer, Bart, Gulcehre, Caglar, Bougares, Fethi,
  Schwenk, Holger, and Bengio, Yoshua.
\newblock Learning phrase representations using rnn encoder-decoder for
  statistical machine translation.
\newblock In \emph{Proceedings of EMNLP}, pp.\  1724--1734, 2014.

\bibitem[Collins et~al.(2005)Collins, Koehn, and
  Ku{\v{c}}erov{\'a}]{collins2005clause}
Collins, Michael, Koehn, Philipp, and Ku{\v{c}}erov{\'a}, Ivona.
\newblock Clause restructuring for statistical machine translation.
\newblock In \emph{Proceedings of ACL}, pp.\  531--540, 2005.

\bibitem[Collobert et~al.(2011)Collobert, Weston, Bottou, Karlen, Kavukcuoglu,
  and Kuksa]{collobert2011natural}
Collobert, Ronan, Weston, Jason, Bottou, L{\'e}on, Karlen, Michael,
  Kavukcuoglu, Koray, and Kuksa, Pavel.
\newblock Natural language processing (almost) from scratch.
\newblock \emph{The Journal of Machine Learning Research}, 12:\penalty0
  2493--2537, 2011.

\bibitem[Graves et~al.(2014)Graves, Wayne, and Danihelka]{ntmgraves2014neural}
Graves, Alex, Wayne, Greg, and Danihelka, Ivo.
\newblock Neural turing machines.
\newblock \emph{arXiv preprint arXiv:1410.5401}, 2014.

\bibitem[Gregor et~al.(2015)Gregor, Danihelka, Graves, and Wierstra]{deepmind}
Gregor, Karol, Danihelka, Ivo, Graves, Alex, and Wierstra, Daan.
\newblock {DRAW}: A recurrent neural network for image generation.
\newblock \emph{arXiv preprint arXiv:1502.04623}, 2015.

\bibitem[Hochreiter \& Schmidhuber(1997)Hochreiter and Schmidhuber]{lstm}
Hochreiter, Sepp and Schmidhuber, J{\"u}rgen.
\newblock Long short-term memory.
\newblock \emph{Neural computation}, 9\penalty0 (8):\penalty0 1735--1780, 1997.

\bibitem[Jean et~al.(2015)Jean, Cho, Memisevic, and
  Bengio]{jean-EtAl:2015:ACL-IJCNLP}
Jean, S\'{e}bastien, Cho, Kyunghyun, Memisevic, Roland, and Bengio, Yoshua.
\newblock On using very large target vocabulary for neural machine translation.
\newblock In \emph{ACL-IJNLP}, 2015.

\bibitem[Kalchbrenner \& Blunsom(2013)Kalchbrenner and
  Blunsom]{kalchbrenner2013}
Kalchbrenner, Nal and Blunsom, Phil.
\newblock Recurrent continuous translation models.
\newblock In \emph{Proceedings of EMNLP}, pp.\  1700--1709, 2013.

\bibitem[Koehn et~al.(2007)Koehn, Hoang, Birch, Callison-Burch, Federico,
  Bertoldi, Cowan, Shen, Moran, Zens, Dyer, Bojar, Constantin, and
  Herbst]{koehn2007}
Koehn, Philipp, Hoang, Hieu, Birch, Alexandra, Callison-Burch, Chris, Federico,
  Marcello, Bertoldi, Nicola, Cowan, Brooke, Shen, Wade, Moran, Christine,
  Zens, Richard, Dyer, Chris, Bojar, Ondrej, Constantin, Alexandra, and Herbst,
  Evan.
\newblock Moses: Open source toolkit for statistical machine translation.
\newblock In \emph{Proceedings of ACL on interactive poster and demonstration
  sessions}, pp.\  177--180, Prague, Czech Republic, June 2007.

\bibitem[Luong et~al.(2015)Luong, Pham, and Manning]{luongEMNLP2015}
Luong, Thang, Pham, Hieu, and Manning, Christopher~D.
\newblock Effective approaches to attention-based neural machine translation.
\newblock In \emph{Proceedings of the 2015 Conference on Empirical Methods in
  Natural Language Processing}, pp.\  1412--1421, 2015.

\bibitem[Pascanu et~al.(2014)Pascanu, Gulcehre, Cho, and
  Bengio]{pascanu2014construct}
Pascanu, Razvan, Gulcehre, Caglar, Cho, Kyunghyun, and Bengio, Yoshua.
\newblock How to construct deep recurrent neural networks.
\newblock In \emph{Proceedings of ICLR}, 2014.

\bibitem[Peng et~al.(2015)Peng, Lu, Li, and Wong]{Reasoning2015}
Peng, Baolin, Lu, Zhengdong, Li, Hang, and Wong, Kam-Fai.
\newblock Towards neural network-based reasoning.
\newblock \emph{arXiv preprint arXiv:1508.05508}, 2015.

\bibitem[Schuster \& Paliwal(1997)Schuster and
  Paliwal]{schuster1997bidirectional}
Schuster, Mike and Paliwal, Kuldip~K.
\newblock Bidirectional recurrent neural networks.
\newblock \emph{Signal Processing, IEEE Transactions on}, 45\penalty0
  (11):\penalty0 2673--2681, 1997.

\bibitem[Stolcke et~al.(2002)]{stolcke2002srilm}
Stolcke, Andreas et~al.
\newblock {SRILM}-an extensible language modeling toolkit.
\newblock In \emph{Proceedings of ICSLP}, volume~2, pp.\  901--904, 2002.

\bibitem[Sutskever et~al.(2014)Sutskever, Vinyals, and Le]{googleS2S}
Sutskever, Ilya, Vinyals, Oriol, and Le, Quoc~VV.
\newblock Sequence to sequence learning with neural networks.
\newblock In \emph{Advances in Neural Information Processing Systems}, pp.\
  3104--3112, 2014.

\bibitem[Vinyals et~al.(2014)Vinyals, Kaiser, Koo, Petrov, Sutskever, and
  Hinton]{vinyals2014grammar}
Vinyals, Oriol, Kaiser, Lukasz, Koo, Terry, Petrov, Slav, Sutskever, Ilya, and
  Hinton, Geoffrey.
\newblock Grammar as a foreign language.
\newblock \emph{arXiv preprint arXiv:1412.7449}, 2014.

\bibitem[Yin et~al.(2015)Yin, Lu, Li, and Ben]{Enquirer2015}
Yin, Pengcheng, Lu, Zhengdong, Li, Hang, and Ben, Kao.
\newblock Neural enquirer: Learning to query tables.
\newblock \emph{arXiv preprint arXiv:1512.00965}, 2015.

\bibitem[Zeiler(2012)]{adadelta}
Zeiler, Matthew~D.
\newblock Adadelta: an adaptive learning rate method.
\newblock \emph{arXiv preprint arXiv:1212.5701}, 2012.

\end{thebibliography}
\bibliographystyle{iclr2016_conference}

\newpage

\section*{APPENDIX:  Actual Translation Examples}
In appendix we give some example translations from \textsc{DeepMemory}, more specically, \textsc{Arc-II} and \textsc{Arc-III}, and compare them against the reference and the translation given by RNNsearch. We will focus on long sentences with relatively complicated structures. \vspace{-5pt}

\subsection*{Example Translation of \textsc{Arc-II}}
\begin{figure}[h!]
\begin{center}
 \includegraphics[width=0.93\textwidth]{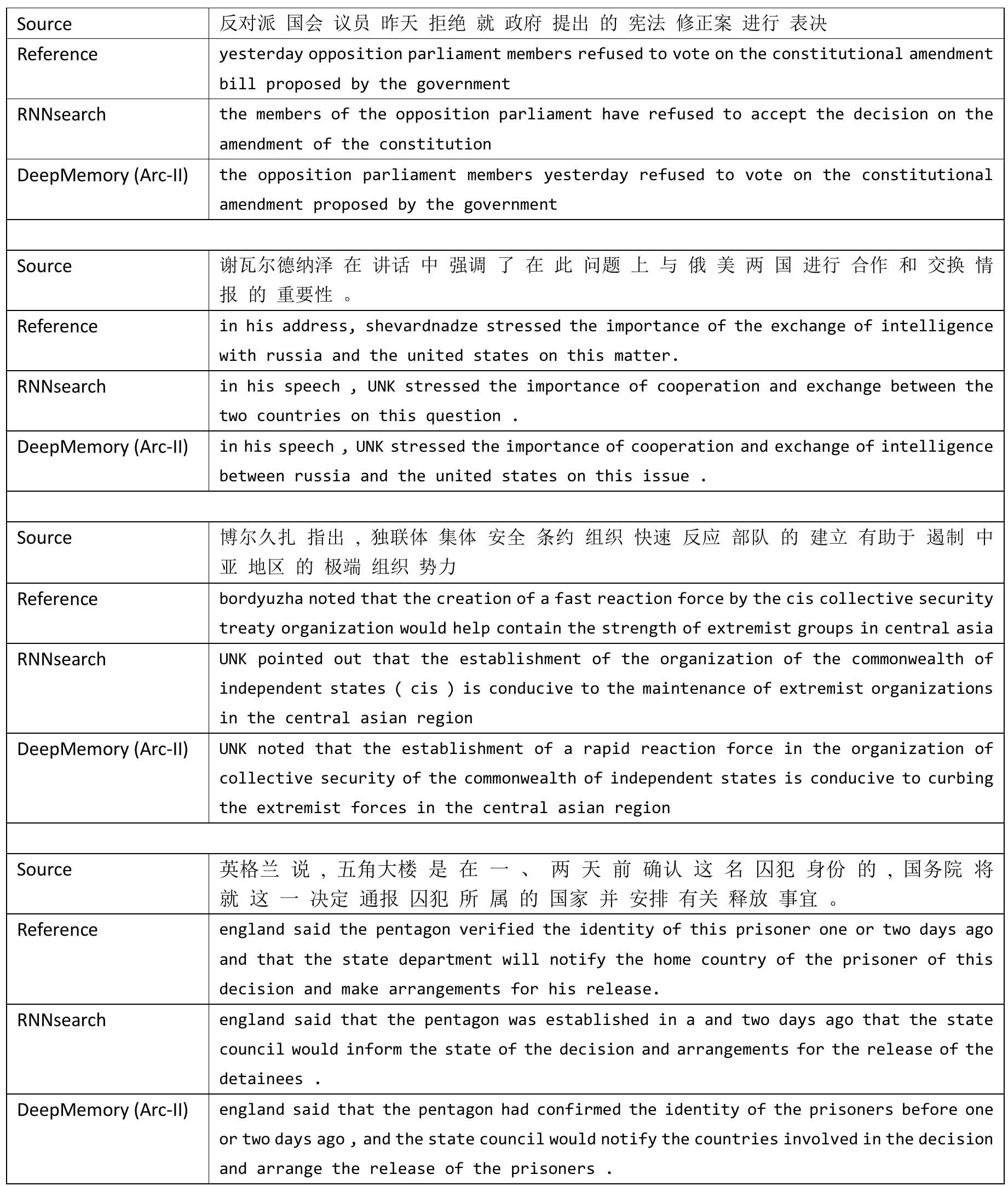}
    \vspace{-12pt}
  \end{center}
\end{figure}
\newpage

\subsection*{Example Translation of \textsc{Arc-III}}
\begin{figure}[h!]
\hspace{2pt}
\begin{center}
 \includegraphics[width=0.93\textwidth]{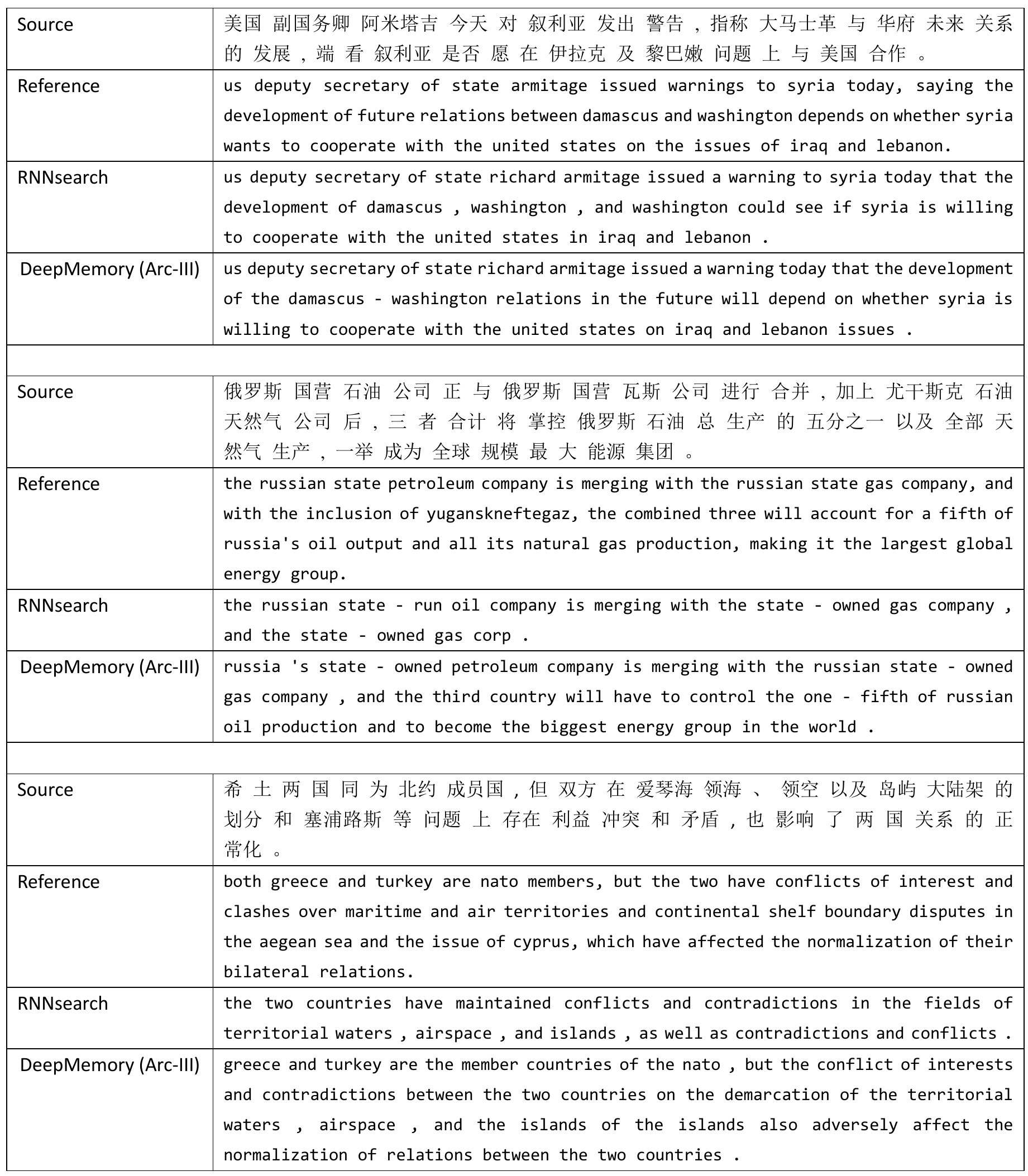}
    \vspace{-12pt}
  \end{center}
\end{figure}

\end{document}